\documentclass[10pt,twocolumn,letterpaper]{article}
\usepackage{graphicx}
\usepackage{caption}
\usepackage{amsmath}

\begin{document}
\title{Megapixel Size Image Creation using Generative Adversarial Networks}
\author{
Marco Marchesi\thanks{marco.marchesi@happyfinish.com}\\
Happy Finish Ltd.
}

\maketitle
\begin{figure*}
  \centering
  \includegraphics[width=\linewidth]{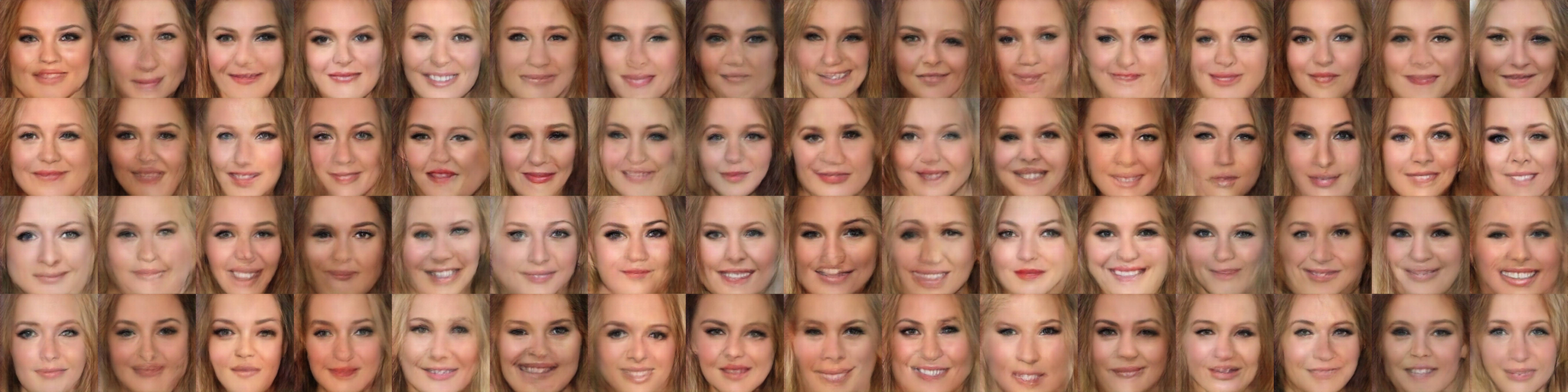}
  \caption{Samples generated at the 256x256 pixels size with the random numbers uniformly distributed in the range [-0.4, 0.4].}
  \label{fig:teaser}
\end{figure*}
\begin{abstract}
Since its appearance, Generative Adversarial Networks (GANs)~\cite{2014arXiv1406.2661G} have received a lot of interest in the AI community. In image generation several projects showed how GANs are able to generate photorealistic images but the results so far didn't look adequate for the quality standard of visual media production industry. We present an optimized image generation process based on a Deep Convolutional Genera\-tive Adversarial Networks (DCGANs), in order to create photorealistic high-resolution images (up to 1024x1024 pixels). Furthermore, the system was fed with a limited dataset of images, less than two thousand images. All these results give more clue about future exploitation of GANs in Computer Graphics and Visual Effects.
\end{abstract}

\section{Introduction}
Generative Adversarial Networks are made by two neural networks competing each other. One, the gene\-rator $G(z)$, it creates images starting from a latent space z of uniformly distributed random numbers, while the discriminator $D(x)$ has to judge the ima\-ges x it receives as fake or real. We train $G(z)$ with the goal to fool $D(x)$ with fake images, minimizing $\log{1 - D(G(z))}$. In order to do that, $G(z)$ has to learn to produce images that are as much photorealistic as possible.\\
This approach is a valid alternative to maximum likelihood techniques, because its conditions and constraints make feasible to run it as an unsupervised learning approach. By the contrary, training is still challenging and efforts are made to prevent both networks to fail.
Several improvements have been introduced since the first GAN model. One of the first techniques was the \textit{minibatch discrimination} that reduces the chance for the generator to collapse~\cite{DBLP:conf/nips/SalimansGZCRCC16}. Other techniques aim to find a faster convergence, modeling the discriminator $D(x)$ as an energy function~\cite{DBLP:journals/corr/ZhaoML16} or introducing new loss definitions~\cite{2017arXiv170310717B}.

\section{The model}
The model we used is a DCGAN~\cite{DBLP:journals/corr/RadfordMC15}, implemented with Google TensorFlow, with a variable batch size depending of the size of the images we wanted to achieve.
For training the discriminator we tested two slightly different datasets (1807 and 1796 images), composed by faces of women taken from magazines and social media. In fact the goal of this project was to generate an image that summarized how the new mums are wrongly represented by media in UK.

\begin{figure}[ht]
  \centering
  \includegraphics[width=3.0in]{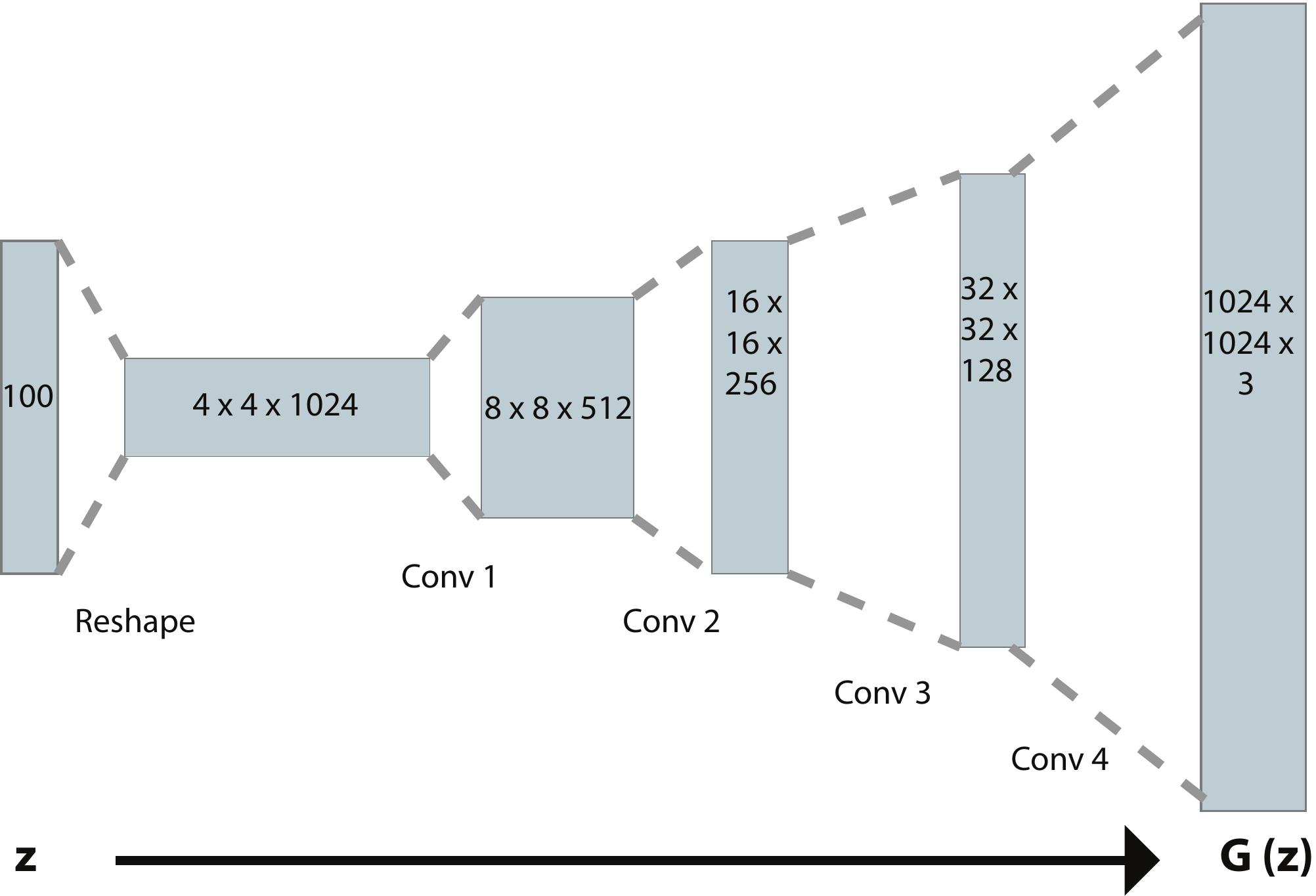}
  \caption{Scheme of $G(z)$ for generating 1024x1024 images.}
  \label{fig:generator}
\end{figure}

For doing that we faced a few challenges:
\begin{itemize}
\item{The dataset was restricted to less than 2k ima\-ges, compared to that ones used on research, thus the system had to learn as much as possible from the limited amount of data. Furthermore, 70\% of the images in the dataset was smaller than 512x512px, so the system had to learn mostly from upscaled images, inferring the high resolution details from the small set of bigger images.}
\item{We trained the system with a NVIDIA Pascal Titan X that was limited in storing a DCGAN able to generate megapixel size images~(Fig.\ref{fig:generator}). For this reason the batch size for the training process was a parameter, starting from 128 (for 192x192px) to 6 (for 1024x1024px).}
\item{ The generated samples had to be photorealistic, to be used commercially, so the system had to limit the artifacts.}
\item{With our dataset, we found that bigger the ima\-ge size, easier for $G(z)$ to diverge.}
\end{itemize}

\section{Training Process}
\begin{figure}[ht]
  \centering
  \includegraphics[width=3in]{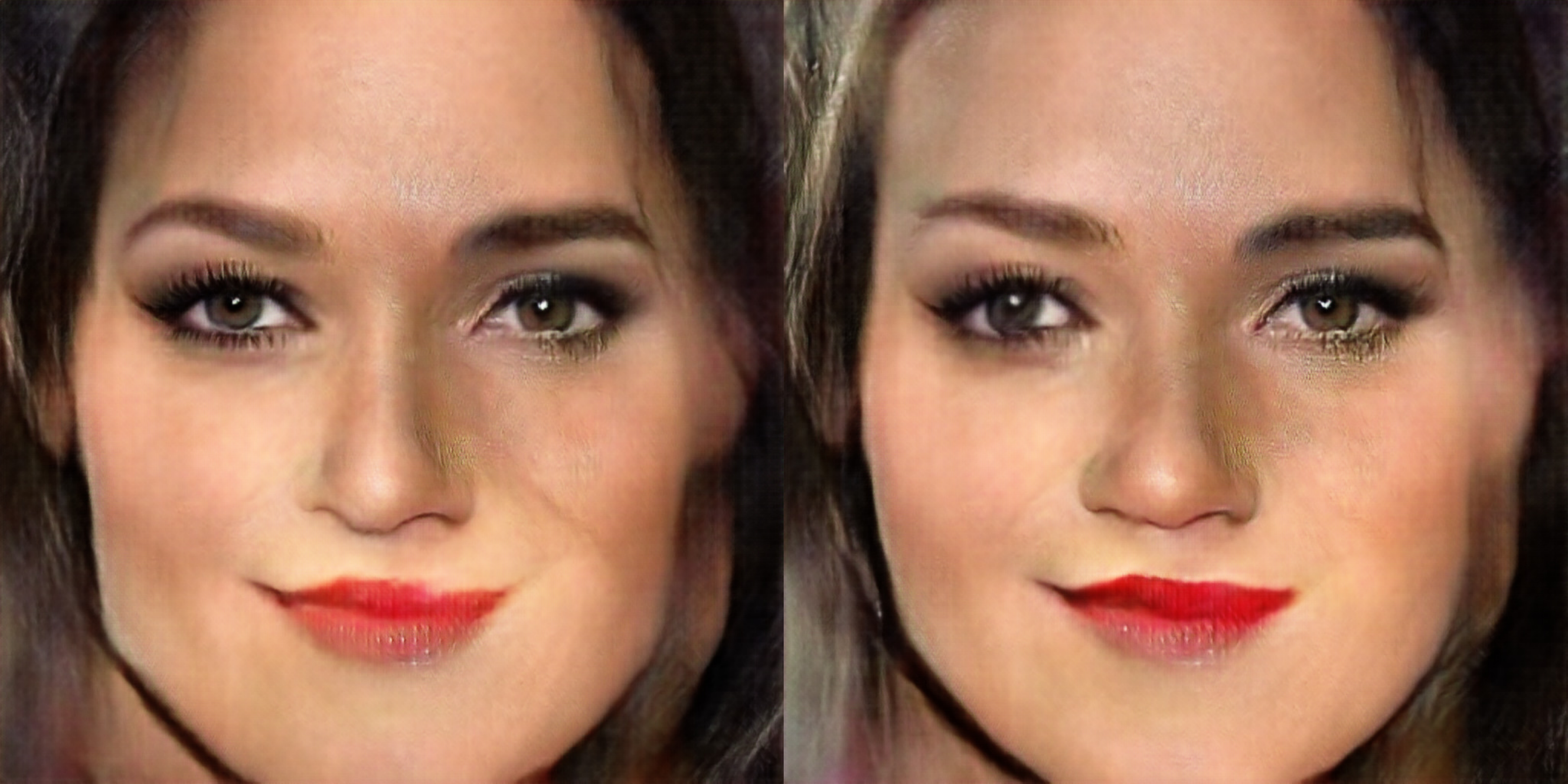}
  \caption{A couple of face variations generated at 1024x1024px.}
  \label{fig:megapixel}
\end{figure}

\begin{figure}[ht]
 \centering
  \includegraphics[width=3in]{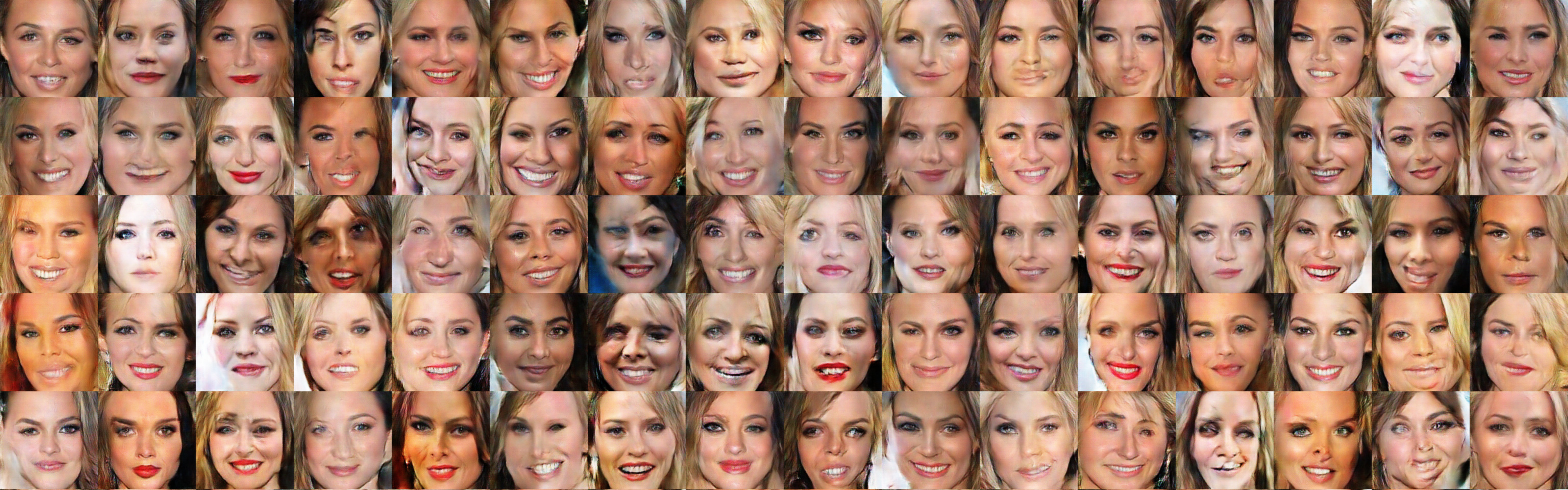}
  \includegraphics[width=3in]{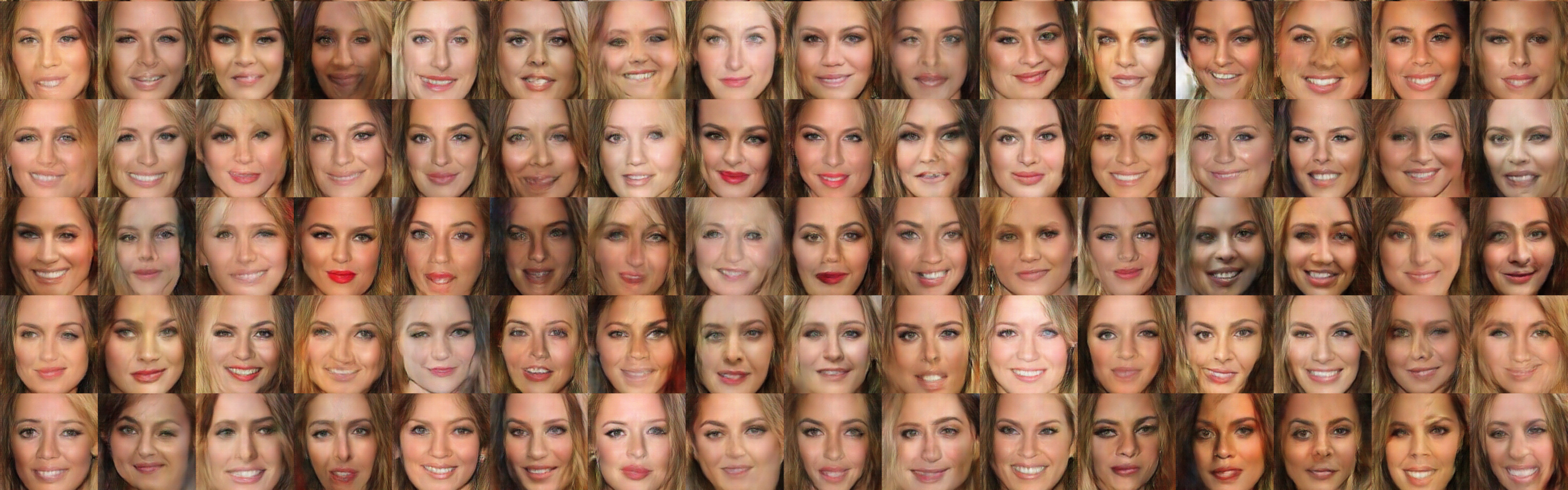}
  \caption{Images generated with a uniform distribution in the intervals $[-1.0, 1.0]$ (\textit{above}) and $[-0.5, 0.5]$ (\textit{below}).}
  \label{fig:after_opt}
\end{figure}

We generated images at different sizes, starting at 192x192px up to 1024x1024px~(Fig.\ref{fig:megapixel}). 
The megapixel size has been produced for the first time, as long as the highest image size for GANs so far was 512px in width~\cite{DBLP:journals/corr/ZhuPIE17}. 
To do so, in brief we applied the following optimizations:
\begin{enumerate}
\item{To prevent the generator and the discriminator to diverge, we applied an additional step for updating alternatively the generator and the discriminator every 50 steps. In this way the loss for both networks oscillated ($loss(D) < 1$ and $loss(G) < 3 $) on a limited interval  but never diverged at any image size.}
\item{For generating the samples, we limited the interval of the uniform distribution of the random inputs $z$. This solution reduced significantly the artifacts, as showed in Fig.\ref{fig:after_opt}.}
\end{enumerate}

\section{Conclusion and Future Work}
We briefly presented the optimization process made on a DCGAN model to generate bigger photorealistic images with a limited dataset. We reached the 1024x1024px size, almost 4x the previous result in research, limiting the artifacts in order to use the image in a creative process for a commercial campaign. We want to test if our improvements can be applied to any dataset. We aim to reduce the memory requirements for GANs, exploiting GPU parallelism, and we want to apply new convergence criteria to GANs, in order to generate even bigger photorealistic images. Further conditional probabili\-ties will let us exploit GANs more widely in other computer graphics fields, like animation and visual effects.

\section*{Acknowledgement}
This research was part of a commercial project funded by MHPC.

\bibliographystyle{ieee}
\bibliography{pm} 
\end{document}